\documentclass{article}
\usepackage{spconf,amsmath,graphicx}
\usepackage{multirow}
\usepackage{makecell}
\usepackage{array}
\usepackage{float}
\usepackage{amsfonts}

\usepackage{booktabs}
\usepackage{xcolor}

\title{Building Robust Spoken Language Understanding by Cross Attention between Phoneme Sequence and ASR Hypothesis}
\name{Zexun Wang$^1$$^{\dagger}$, Yuquan Le$^{1,2}$$^{\dagger}$\thanks{$^{\dagger}$ Equal contribution.}, Yi Zhu$^{1,3}$, Yuming Zhao$^1$, Mingchao Feng$^1$, Meng Chen$^1$, Xiaodong He$^1$}
\address{
  $^1$JD AI, Beijing, China, $^2$Hunan University, $^3$LTL, University of Cambridge\\
  \texttt{leyuquan@hnu.edu.cn}, \texttt{yz568@cam.ac.uk}\\
  \texttt{\{wangzexun3,zhaoyuming3,fengmingchao,chenmeng20,xiaodong.he\}@jd.com}
  }
\begin{document}
%\ninept
%
\maketitle
\begin{abstract}
Building Spoken Language Understanding (SLU) robust to Automatic Speech Recognition (ASR) errors is an essential issue for various voice-enabled virtual assistants. Considering that most ASR errors are caused by phonetic confusion between similar-sounding expressions, intuitively, leveraging the phoneme sequence of speech can complement ASR hypothesis and enhance the robustness of SLU. This paper proposes a novel model with \textbf{C}ross \textbf{A}ttention for \textbf{SLU} (denoted as \textbf{CASLU}). The cross attention block is devised to catch the fine-grained interactions between phoneme and word embeddings in order to make the joint representations catch the phonetic and semantic features of input simultaneously and for overcoming the ASR errors in downstream natural language understanding (NLU) tasks.
%to fuse the information from phoneme sequence and ASR hypothesis seamlessly. The cross attention block is devised to catch the fine-grained interactions between phoneme embeddings and word representations, then lexical-aware phoneme representations and phonetic-aware text representations are obtained based on attention weights. Finally, the fused joint representations catch the phonetic and semantic features of input simultaneously and are capable of overcoming the ASR errors in downstream NLU tasks. 
% Extensive experiments are conducted on three datasets. We also validate the universality of our model and prove the complementarity of our model when combining with other robust SLU techniques. Experimental results show the effectiveness and competitiveness of our approach. 
Extensive experiments are conducted on three datasets, showing the effectiveness and competitiveness of our approach.
Additionally, We also validate the universality of CASLU and prove its complementarity when combining with other robust SLU techniques.  

\end{abstract}
\section{Introduction}
Spoken Language Understanding (SLU) is the critical technology of voice-enabled virtual assistants, e.g. Apple Siri and Amazon Alexa. It serves as a bridge to allow machines to interact with humans effectively and has obtained increasing attention in recent years. The SLU generally involves two main modules, Automatic Speech Recognition (ASR) and Natural Language Understanding (NLU). The ASR engine is utilized to transcribe human speech into the text. Then the NLU module is applied to ASR output to comprehend the user's requests, typically including intent classification and slot filling tasks. Owing to the remarkable success in various fields, deep learning techniques are widely explored for SLU~\cite{yao2014spoken,goo2018slot,mesnil2014using}. %\cite{yao2014spoken,guo2014joint,mesnil2014using,goo2018slot}. 
% Owing to the remarkable success in various fields, e.g. natural language processing and computer vision, deep learning techniques are also explored for SLU~\cite{yao2014spoken,goo2018slot}.
% Although these methods have made rapid progress, the performance is still compromised inevitably when facing ASR errors \cite{ogawa2017error}. A common ASR error is that the similar-sounding words are incorrectly transcribed to each other. For example, ``\textit{buy a computer}" may be mis-recognized to ``\textit{by a computer}", which would confuse the downstream task. Hence, building SLU robust to ASR errors is essential for improving the end-user experience in voice-based virtual assistants.
Although these methods have made rapid progress, the performance is still compromised inevitably when facing ASR errors \cite{ogawa2017error}. Hence, building SLU robust to ASR errors is essential for improving end-user experience in virtual assistants.%voice-based virtual assistants.

% Previous works related to robust SLU mainly focus on leveraging multiple ASR outputs, e.g. n-best hypotheses, to recover the correct transcriptions~\cite{li2020multi,liu2021asr,ruan2020towards}.
% %~\cite{li2020multi,li2020improving,liu2021asr,ruan2020towards}.
% However, as shown previously, many ASR errors are caused by phonetic confusion, so it is difficult to fully address this problem only with text-level information which does not carry any sound information.
% %Therefore, in this paper, we propose to harness phoneme-level information to complement ASR transcriptions for more robust SLU.
% As the atom sound unit of a language, phoneme captures complex phonetic properties and interactions of speech, so the ASR transcription in phoneme level should be more similar to the correct utterance than character or word level \cite{sundararaman2021phoneme}, and the Phoneme Error Rate (PER) will be smaller than Character Error Rate (CER) and Word Error Rate (WER). 
% Therefore, in this paper, we propose a novel deep learning-based model CASLU to harness phoneme-level information to complement ASR transcriptions for more robust SLU. The lower-level (phoneme) representations catch the phonetic features of speech, which can alleviate the ASR errors. The higher-level (character or word) representations cover the semantic meanings, which are more conducive to downstream NLU tasks. We argue that effectively fusing the two kinds of complementary information can improve the robustness of SLU to ASR errors. To the best of our knowledge, there has been rare work exploring this before.

Previous works related to robust SLU mainly deal with ASR transcripts directly. 
Various approaches have been investigated to correct ASR hypothesis~\cite{DBLP:conf/icassp/TamLZW14,weng2020joint}, or leverage ASR output information~\cite{ladhak2016latticernn,huangchen2020learning,tur2013semantic,shivakumar2019spoken}, especially N-best hypothesis~\cite{li2020improving,liu2021asr,ogawa2018rescoring,ogawa2019improved} directly within downstream NLU model.
Despite the success of these methods, using only texts generated by ASR module unavoidably loses useful speech information like pronunciation and prosody. 
Moreover, many ASR errors are even elicited by phonetic confusion of similar-sounding words that are incorrectly transcribed to each other.
%\yicomment{any reference here?} 
For example, ``\textit{buy a computer}" may be mis-recognized to ``\textit{by a computer}", which would confuse the downstream task.
To this end, there have been research efforts drawing upon speech information exploiting different forms to improve SLU robustness~\cite{ruan2020towards,serdyuk2018towards,lugosch2019speech,fang2020using,sundararaman2021phoneme,9414558}.
Among them, phoneme is considered as a clean form\footnote{For example, audio signals or features may vary greatly across people and could be distorted by different noise sources.} of speech representation complementary to text. 
As the atom speech unit of a language, phoneme can capture complex phonetic properties and interactions, so the ASR transcription in phoneme level should be more similar to the correct utterance than character or word level \cite{sundararaman2021phoneme}, and the Phoneme Error Rate (PER) will be smaller than Character Error Rate (CER) and Word Error Rate (WER). 

% capable of capturing complex phonetic properties and interactions.

Inspired by this, we propose a novel deep learning-based model CASLU to harness phoneme-level information to complement text for more robust SLU. 
%In particular, we focus on fusing the information of ASR hypothesis and phoneme sequence by catching their fine-grained interactions. The lower-level (phoneme) representations catch the phonetic features of speech, which can alleviate the ASR errors. The higher-level (character or word) representations cover the semantic meanings, which are more conducive to downstream NLU tasks. These two types of information complement each other and are both critical for the SLU system. 
In particular, we utilize cross-attention~\cite{hou2019cross} to explicitly model the fine-grained interactions between ASR phonemes and hypotheses.
To the best of our knowledge, there has been rare work exploring this before. 
%only with a concurrent work~\cite{sundararaman2021phoneme} using a BERT-based model~\cite{DBLP:conf/naacl/DevlinCLT19} to train on ASR transcript and phoneme sequence jointly.
The main contributions of this paper are two-fold:
%\begin{itemize}
(1) We propose CASLU to explicitly capture the fine-grained correlations between ASR phonemes and hypotheses for robust SLU. (2) Experimental results on three datasets demonstrate the effectiveness and competitiveness of CASLU. We also validate the universality of CASLU with different text and phoneme encoders, and prove its complementarity combining with other robust SLU techniques. 
\begin{figure}[!t]
\begin{center}
\includegraphics[width=0.6\columnwidth, trim={9.2cm 1.3cm 12.9cm 0.8cm}, clip]{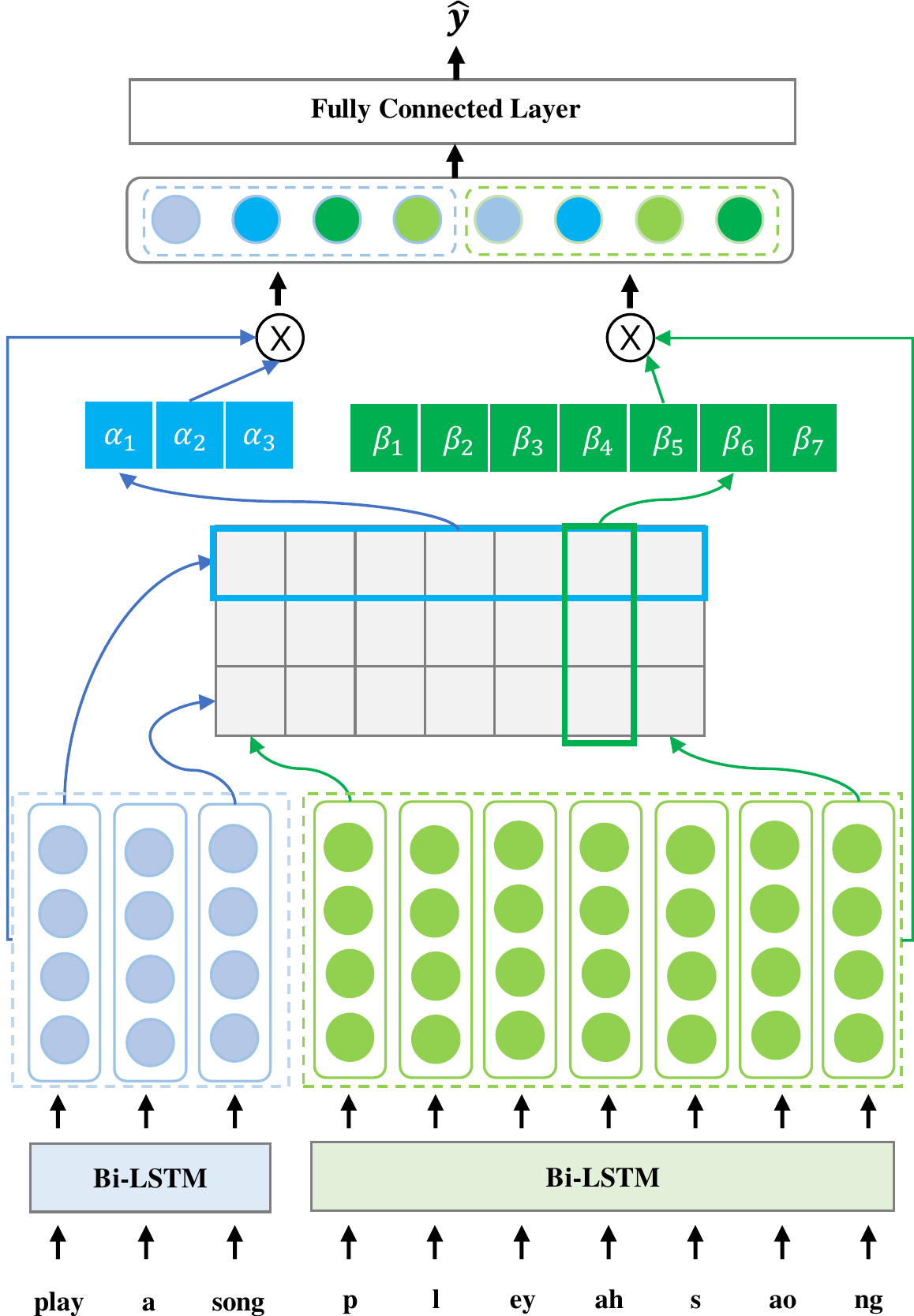}
\vspace{-2mm}
\caption{\rm \centering The architecture of our proposed model CASLU.}
\label{fig:model}
\end{center}
\vspace{-5mm}
\end{figure}
\vspace{-2mm}
\section{Model}

This section introduces our proposed model CASLU in detail (see Figure \ref{fig:model}). The input is an utterance X which contains a sequence of words as well as its corresponding phoneme sequence: $X^w=[x^w_1,x^w_2,...,x^w_m]$ and $X^p = [x^p_1, x^p_2, ..., x^p_n]$, where $x^w_i \in V_w$, $V_w$ is the word vocabulary, $m$ is the sequence length of text sequence, $x^p_j \in V_p$, $V_p$ is the phoneme vocabulary, and $n$ is the length of the phoneme sequence.
The goal of the model is to interpret the user's utterance request to its corresponding class $y$.

\subsection{Text and phoneme Encoder}
The text encoder utilizes Bi-LSTM~\cite{hochreiter1997long} to extract word information in context. %the Bi-directional Long Short-Term Memory
Each word $x^w_i$ in $X^w$ is mapped into its corresponding word vector $\mathbf{e}^w_i \in \mathbb{R}^{d_w}$, where $d_w$ is the dimensionality of word embedding. Then a Bi-LSTM encoder computes contextualized hidden representations as follows:
%$X^w=[x^w_1,x^w_2,...,x^w_m]$, each word $x^w_i$ is mapped into its corresponding word vector $\mathbf{e}^w_i \in \mathbb{R}^{d_w}$, where $d_w$ is the dimensionality of word embedding. Then the text encoder layer uses Bi-LSTM to obtain contextualized word hidden representations. The formula is as follows:
\vspace{-1mm}
\begin{equation}
\label{eq:bi_lstm2word}
\resizebox{0.90\linewidth}{!}{
  \begin{minipage}{\linewidth}
$
\begin{aligned}
    %  & \mathbf{e}_i^w=\mathbf{W}_w\mathbf{x}_i^w, i \in \{1, 2, ..., m\} \\
    %  & \overrightarrow{\mathbf{h}_i^w} = \overrightarrow{\mathbf{LSTM}}(\mathbf{e}_i^w), i \in  [1,m] \\
    %  & \overleftarrow{\mathbf{h}_i^w} = \overleftarrow{\mathbf{LSTM}}(\mathbf{e}_i^w), i \in [m,1] \\
    %  & \mathbf{h}_i^w=[\stackrel{\rightarrow}{\mathbf{h}_i^w}, \stackrel{\leftarrow}{\mathbf{h}_i^w}]
     & \overrightarrow{\mathbf{h}_i^w} = \overrightarrow{\mathbf{LSTM}}(\mathbf{e}_i^w), \;
     \overleftarrow{\mathbf{h}_i^w} = \overleftarrow{\mathbf{LSTM}}(\mathbf{e}_i^w), \;
     \mathbf{h}_i^w=[\overrightarrow{\mathbf{h}_i^w}, \overleftarrow{\mathbf{h}_i^w}]
\end{aligned}
$
\end{minipage}}
\vspace{-1mm}
\end{equation}
where $\overrightarrow{\mathbf{h}_i^w}$ and $\overleftarrow{\mathbf{h}_i^w}$ are the forward and backward hidden states for word $x_i^w$. $\mathbf{h}_i^w$, the concatenation of hidden states in both directions, is used as the word hidden representations.

Similar to text encoder, the phoneme sequence is fed into phoneme encoder with another Bi-LSTM to acquire contextualized phoneme hidden representations $\mathbf{h}_j^p$
for each phoneme.
The process is similar to Equation \ref{eq:bi_lstm2word}.

\begin{table}[!t]
\centering
\renewcommand\arraystretch{0.92}
\begin{tabular}{lcccc}
\toprule
Datasets&Train&Dev&Test&Intents\\\midrule
Snips&11769&1312&700&7\\
%\midrule  %添加表格中横线
TREC&4904&548&500&6\\
%\midrule  %添加表格中横线
Waihu&7141&906&886&16\\
\bottomrule
% \hline %添加表格底部粗线
\end{tabular}
\caption{\rm \centering Dataset statistics.}
\label{table:dataset}
\vspace{-3mm}
\end{table}

\subsection{Interaction Layer}
After encoding text and phoneme sequences, it is now crucial to fuse their representations in order to effectively leverage information from both sides.
%Recently, the cross attention network has been proposed \cite{hou2019cross} and applied in the few-shot image classification task. 
The cross attention block has been proved to be capable of capturing fine-grained correlation of any pixel pair between two images \cite{hou2019cross}. 
%This advantage would be quite appealing for our task. 
Motivated by this, 
%we devise a novel cross attention block to seamlessly exploit word-phoneme pair correlation and make comprehensive interaction. In particular, 
we first calculate a correlation map $\mathbf{C} \in \mathbb{R}^{m \times n}$ between word hidden representations $\mathbf{h}^w_i$ and phoneme hidden representations $\mathbf{h}^p_j$ through cosine distance: %The detail is as follows:
\begin{equation}
\begin{array}{ll}
C_{ij} = \frac{(\mathbf{h}^w_i)^T \mathbf{h}^p_j}{||\mathbf{h}^w_i||_2  ||\mathbf{h}^p_j||_2}
\end{array}
\label{eq:correlation_map}
\end{equation}
where $C_{ij}$ can be seen as the semantic relevance between word $x^w_i$ and phoneme $x^p_j$, and the matrix $\mathbf{C}$ characterizes the fine-grained correlations among words and phonemes. Specifically, the vector of the $i$-th row in $\mathbf{C}$: 
\begin{equation*}
\begin{aligned}
&\mathbf{r}_{i} = 
\begin{bmatrix}
C_{i, 1} & C_{i, 2} & ... & C_{i, n}
\end{bmatrix}^T, \mathbf{r}_i \in \mathbb{R}^{n} \textrm{ and } i \in \{1, ..., m\}
\end{aligned}
\end{equation*}
represents the correlations between the $i$-th word $x^w_i$ and \textit{every} phoneme in the phoneme sequence $X^p$. Correspondingly, the $j$-th column vector $\mathbf{c}_j \in \mathbb{R}^{m}$ denotes the relationship between the $j$-th phoneme $x^p_j$ and the whole word sequence $X^w$.

We further apply convolution operation to \textit{each} row and column vector of $\mathbf{C}$ individually to fuse local correlations between words and phonemes into cross attention weight.
Taking the row vector $\mathbf{r}_i$ as an example, which represents interactions of the word $x^w_i$ over phoneme sequence, its attention weight $\alpha_i$ is calculated by employing a convolution layer with a single text kernel $\mathbf{k}^w \in \mathbb{R}^n$ followed by softmax function:
\begin{equation}
\label{eq:word_attention}
\begin{aligned}
& \alpha_i = \frac{\exp({\mathbf{k}^w}^T \mathbf{r}_i)}{\sum_{i'=1}^{m}\exp({\mathbf{k}^w}^T\mathbf{r}_{i'})}, i \in \{1, ..., m\}
\end{aligned}
\end{equation}
Similarly, a phoneme kernel $\mathbf{k}^p \in \mathbb{R}^m$ operates on column vectors $\mathbf{c}_j$ to obtain the attentions $\beta_j$ for all input phonemes.\footnote{In practice, the length $m$ and $n$ of kernels $\mathbf{k}^w$ and $\mathbf{k}^p$ are the maximum lengths of all text and phoneme sequences.}

% $n \times 1$ kernel ($\mathbf{w} \in R^{n \times 1}$) to fuse each local correlation vector $\mathbf{c}^w_i$ of $\mathbf{C}$ into an attention scalar. On the other hand, the $j$-th column $\mathbf{c}^p_j \in R^{m \times 1}$ denotes the relationship among the $j$-th phoneme $x^p_i$ and the word sequence $X^w$. Similarly, we apply convolution operation with a $m \times 1$ kernel ($\mathbf{w}^{'} \in R^{m \times 1}$) to fuse each local correlation vector $\mathbf{c}^p_j$ of $\mathbf{C}$ into an attention scalar. By this way, we can generate the word attention vector and phoneme attention vector. Taking the word attention vector as an example, the word attention vector $\alpha \in R^{m \times 1}$ is calculated as follows:

Finally, after aggregating the feature map $\mathbf{C}$ into attentions for all words and phonemes with convolution kernels, the \textit{phonetic-aware text representation} $\mathbf{t}$ and \textit{lexical-aware phoneme representation} $\mathbf{p}$ can be computed as follows:
\begin{equation}
\label{eq:final_embed}
\begin{aligned}
& \mathbf{t} = \mathbf{h}^w\boldsymbol{\alpha}, \quad \mathbf{p} = \mathbf{h}^p\boldsymbol{\beta}
\end{aligned}
\end{equation}
where $\mathbf{h}^w = [\mathbf{h}^w_1, ..., \mathbf{h}^w_m]$ stacks all the word hidden representations and $\boldsymbol{\alpha} = [\alpha_1, ..., \alpha_m]^T$, and $\mathbf{h}^p$ and $\boldsymbol{\beta}$ are defined analogously.

\begin{table*}[!t]
\centering
% \caption{\rm \centering Performance comparison in accuracy of NLU models. Trans and ASR mean evaluating on transcription $x_{trs}$ and 1-best ASR hypothesis $x_{asr}$ respectively.}
\begin{tabular}{lcccccc}
\hline
\multirow{2}*{Models} & \multicolumn{2}{c}{Snips} & \multicolumn{2}{c}{TREC} & \multicolumn{2}{c}{Waihu}\\
&Trans(\%)& ASR(\%)& Trans(\%)& ASR(\%)& Trans(\%)& ASR(\%)\\
\hline  %添加表格中横线
%Bi-LSTM(train w/trs)& 96.24& 73.10& 84.13& 62.27& 72.16&67.95\\
%Bi-LSTM(train w/asr) & 94.57& 90.78& 83.40& 69.40& 70.09& 68.47\\
B1: Bi-LSTM w/ trs & \textbf{96.24}& 73.10& 84.13& 62.27& 72.16&67.95\\
B2: Bi-LSTM w/ asr & 94.57& 90.78& 83.40& 69.40& 70.09& 68.47\\
%Conf-aware FT& \textbf{96.99}&	92.41& \textbf{88.46}& 72.22& /& /\\
B3: Multi-input \cite{fang2020using} & 94.50& 92.00& 83.60& 72.00& 70.69& 68.89\\
\hline
CASLU& 95.50& \textbf{92.57}& \textbf{84.47}& \textbf{73.60}& \textbf{73.55}& \textbf{69.22}\\
CASLU w/o $\mathbf{t}$ & 92.14& 91.63& 82.60& 71.35& 70.15& 68.79\\
CASLU w/o $\mathbf{p}$ & 94.71& 91.43& 81.40& 71.20& 73.13& 68.62\\
\hline %添加表格底部粗线
\end{tabular}
\label{table:main_result}
\caption{Performance comparison in accuracy of CASLU and other baselines. 
\textit{Trans} and \textit{ASR} mean evaluating on transcription $x_{\textrm{trs}}$ and 1-best ASR hypothesis $x_{\textrm{asr}}$ respectively.
\textit{CASLU w/o $\mathbf{t}$ ($\mathbf{p}$)} denotes omitting $\mathbf{t}$ ($\mathbf{p}$) in the final prediction layer.}
\vspace{-4mm}
\end{table*}

% The kernel $w$ aggregates the correlations between the word $x^w_i$ and all phoneme sequence ${X^p}$ as the attention scalar. Then, the \textit{phonetic-aware text representation} $\mathbf{r}^{pa}$ can be obtained as follows:

% \begin{equation}
% \label{eq:phonetic_aware_text_embed}
% \begin{aligned}
% & \mathbf{r}^{pa} = \alpha^T \mathbf{h}^w  
% \end{aligned}
% \end{equation}
% Similarly, we can obtain the phoneme attention vector $\beta \in R^{n \in 1}$. Then the \textit{lexical-aware phoneme representation} $\mathbf{r}^{la}$ can be obtained as follows:

% \begin{equation}
% \label{eq:lexical_aware_phoneme_embed}
% \begin{aligned}
% & \mathbf{r}^{la} = \beta^T \mathbf{h}^p  
% \end{aligned}
% \end{equation}

%\subsection{Prediction Layer}
% \subsection{Prediction}
%The outputs of the interaction layer are phonetic-aware text representation $\mathbf{t}$ and lexical-aware phoneme representation $\mathbf{p}$. 
We concatenate $\mathbf{t}$ and $\mathbf{p}$ and feed the resulted vector into a fully connected layer with \textit{softmax} function to predict the probability of intent class. The standard cross-entropy is applied to calculate the classification loss. 
%The detailed formulas are as follows:

%Hereafter, the output of fully connected layer is passed to a softmax function to obtain the possibility of intent prediction. The cross-entropy with Adam optimizer is applied to optimize the objective. The detailed formulas are as follows:
%These two kinds of information complement each other and thus are crucial for tackling the ASR errors in downstream NLU tasks. Therefore, we concatenate them and feed the resulted vector into a fully connected layer with softmax function to predict the probability of intent class. The standard cross-entropy is applied as the classification loss. The detailed formulas are as follows:
%Hereafter, the output of fully connected layer is passed to a softmax function to obtain the possibility of intent prediction. The cross-entropy with Adam optimizer is applied to optimize the objective. The detailed formulas are as follows:

% \begin{equation}
% \label{eq:concat}
% \begin{aligned}
% & \mathbf{h^o} = [\mathbf{t},\mathbf{p}]\\
% & \mathbf{\hat{y}} = \textrm{softmax}(\mathbf{W}_o \mathbf{h}^o + \mathbf{b})\\
% & \mathcal{L} = \sum_{i\in D}^{} \textbf{CE}(\mathbf{y}_i, \mathbf{\hat{y}_i})
% \end{aligned}
% \end{equation}
% where $\mathbf{W}_o$ is the weight matrix, and $\mathbf{b}$ is the bias. 
% $D$ is the number of training samples, $\textbf{CE(·, ·)}$ is the cross-entropy function, and $\mathbf{\hat{y}}_i$ and $\mathbf{y}_i$ are the predicted label distribution and the ground-truth one-hot label, respectively. 

\section{Experiments}

\subsection{Dataset}
\label{sec:dataset}
Experiments are conducted on three datasets, and their statistics are shown in
Table~\ref{table:dataset}: (1) \textbf{Snips} is a benchmark dataset for SLU intent detection. It comprises pairs of user commands and intents such as GetWeather and PlayMusic. We held out a validation set from the training set. (2) \textbf{TREC} is a question classification dataset containing six fact-based question types such as HUMAN and LOCATION. (3) \textbf{Waihu} is a Chinese dataset for intent classification, which was collected from our online voice-enabled customer service bot,\footnote{http://yanxi.jd.com.} so the audio is from real users and the ASR hypotheses and phoneme sequences are generated by our online ASR engine.  
As for Snips and TREC with only text, since speech transcription and annotation are expensive and labor-intensive, we follow a similar strategy as described in \cite{fang2020using} to create noisy speech corpus form them. 
We use Amazon Polly\footnote{https://aws.amazon.com/cn/polly.} to convert the raw text to speech and apply Speech Synthesis Markup Language (SSML) tags and ambient noise\footnote{www.pacdv.com/sounds/ambience\_sounds.html.} to get noisy speech data. Then, we use Amazon Transcribe\footnote{https://aws.amazon.com/transcribe.} to transcribe the audio. Since our hypothesis is to create a  model that is robust to ASR errors, we keep only hypotheses containing ASR errors. Therefore, the phoneme sequences from ASR systems should link the phonemes that the system often confuses.

\subsection{Training Details}

For all datasets, the maximum length of text and phoneme sequence is set to 40 and 80 respectively. Masks are applied to zero out the effect of paddings. 
%For Waihu dataset, all the utterances are processed by jieba\footnote{https://github.com/fxsjy/jieba.} for word segmentation. 
The model uses one layer of Bi-LSTM with 150 hidden nodes. The embeddings of word and phoneme are randomly initialized for simplicity. %\footnote{Using pre-trained word embeddings theoretically benefits all models.} 
The batch-size is set to 64. Adam~\cite{DBLP:journals/corr/KingmaB14} is used as the optimizer with a learning rate of 0.001. We train the model for 20 epochs and choose the best model on the validation set. 
%For all baselines and our proposed model, we use one layer of Bi-LSTM with 150 hidden nodes. The word embeddings and phoneme embeddings are randomly initialized for simplicity.\footnote{Using pre-trained word embeddings theoretically benefits all models.} The batch-size is set to 64. Adam~\cite{DBLP:journals/corr/KingmaB14} is used as the optimizer with a learning rate of 0.001. We train the model for 20 epochs and choose the best model on the validation set. 
We run three trials for each experiment with different random seeds and report the average score to avoid bias introduced by training randomness.

% Considering that deep neural networks training is a stochastic process, we run multiple trials for each experiment and an average of multiple trials is reported to avoid bias introduced by randomness.

\subsection{Main results}
For evaluation, we use classification \textit{accuracy} as metric.
We test on both manual transcriptions $x_{\textrm{trs}}$ and ASR hypotheses $x_{\textrm{asr}}$ on account of the criterion from~\cite{ruan2020towards} that \textit{model robustness towards ASR errors is only improved given increased performance tested on a set with ASR errors and no performance degradation on a set without ASR errors}. Experimental results are reported in Table 2. 
The \textbf{B1} and \textbf{B2} are the baselines that use \textit{only} manual transcriptions (clean text) and 1-best ASR hypotheses (noisy text) as input respectively. The  \textbf{B3} concatenates $\mathbf{t}$ and $\mathbf{p}$ averaged over $\mathbf{h}^w$ and $\mathbf{h}^p$ rather than the weighted sum in Equation~\ref{eq:final_embed}. It's a special case of cross-attention where attention weights are just uniform.
In the first part of Table 2, we observed that: (1) CASLU comprehensively beats all baselines on the test sets of ASR hypotheses (p\textless0.05); (2) When evaluating on manual transcriptions, CASLU even performs better on TREC and Waihu with only a slight drop on Snips. Both results indicate our model enhanced by phoneme sequence can achieve stronger robustness to ASR errors.
%Both results are consistent with the previous evaluation criterion in~\cite{ruan2020towards}, indicating our model enhanced by phoneme sequence can achieve stronger robustness to ASR errors.
%can also boost the accuracy on clean text and proved the robustness of CASLU method. 
The second part shows the ablation study. It has the same architecture as the full CASLU, except that it only feeds the representation of phoneme/text into the final classifier layer. After removing either text or phoneme representation, the performance degrades in both scenarios, which suggests that representations in both modalities complement each other. However, it is noteworthy that CASLU w/o $\mathbf{t}$ or CASLU w/o $\mathbf{p}$ are both better than B2 on ASR hypotheses, which means that after fine-grained interaction, either phonetic-aware text representation or lexical-aware phoneme representation has fused the information from the other.

\subsection{Discussion}
\label{dis}
%This section lists five research questions that guide the discussion: \textbf{RQ1}: Does our method have universality for other popular deep learning models? \textbf{RQ2}: Can our method be integrated with other methods? Whether it can improve? \textbf{RQ3}: Which situations can be alleviated by introducing phonemes? Can we do a case analysis? \textbf{RQ4}: How to interpret the correlation map? Can we visualize it? \textbf{RQ5}: Can we analyze the advantages and disadvantages of different methods?
%This section lists four research questions that guide the discussion.

\noindent \textbf{Universality of our method.} 
To validate the effectiveness of CASLU on other neural structures, here we implement other variant models by replacing Bi-LSTM encoder with GRU/LSTM/Bi-GRU/CNN. We keep the model structure and hyper-parameters consistent for fair comparison. For simplicity, we perform experiments on Snips and Waihu datasets. Table \ref{table:university} shows that our proposed models still outperform the corresponding baselines, which demonstrates that our method can be a unified framework for SLU. 
%In CNN encoder, compared to Multi-input, the effect of CASLU does not seem as significant as in other encoders. This could possibly be attributed to the CNN convolutions that obtain the local field of sequence. The convoluted text/phoneme representations before cross attention may not correspond to each specific word/phoneme token, but tokens received by a local filter.

% of word or phoneme obtained after convolution may not be the meaning of a word or a phoneme itself, but the meaning of words or phonemes in a local filter

% CASLU+VAT is the integration of CASLU and VAT, and CASLU+N-best uses N-best ASR hypothesis texts and phonemes as input.

\begin{table}[!t]
\centering
\footnotesize
\begin{tabular}{llcccc}
\toprule
% \multirow{2}*{Models} & \multicolumn{2}{c}{Snips}  & \multicolumn{2}{c}{Waihu}\\\cmidrule(lr){2-3}\cmidrule(lr){4-5}
&& \multicolumn{2}{c}{Snips}  & \multicolumn{2}{c}{Waihu}\\\cmidrule(lr){3-4}\cmidrule(lr){5-6}
&&Trans(\%)& ASR(\%)& Trans(\%)& ASR(\%)\\
\hline  %添加表格中横线
\multirow{3}{*}{\rotatebox[origin=c]{90}{GRU}} & w/ asr & 94.21& 90.31& 65.46& 63.66\\
& Multi-input & 94.53& 90.50& 67.23& 65.35\\
& CASLU & \textbf{95.17}& \textbf{91.55}& \textbf{70.62}& \textbf{67.47}\\
\hline  %添加表格中横线
\multirow{3}{*}{\rotatebox[origin=c]{90}{LSTM}} & w/ asr& 94.86& 89.84& 65.23& 62.53\\
& Multi-input & 95.14& 90.67& 66.76& 64.80\\
& CASLU & \textbf{95.35}& \textbf{91.26}& \textbf{68.84}& \textbf{66.89}\\
\hline  %添加表格中横线
\multirow{3}{*}{\rotatebox[origin=c]{90}{Bi-GRU}} &
w/ asr& 94.43& 91.33& 70.03& 68.28\\
& Multi-input & 95.10& 91.71& 72.80& 69.47\\
& CASLU & \textbf{95.24}& \textbf{92.38}& \textbf{73.36}& \textbf{70.71}\\
\hline
\multirow{3}{*}{\rotatebox[origin=c]{90}{CNN}} & w/ asr &95.29 &91.95 &73.55& 70.13\\
& Multi-input & 95.95& 92.10&73.65&70.81 \\
& CASLU & \textbf{96.29}& \textbf{92.57}& \textbf{73.97}& \textbf{70.88}\\
\bottomrule
%\hline{1} %添加表格底部粗线
\end{tabular}
\caption{Test accuracy of variant models with other encoder architectures, shown in the leftmost column.}
\label{table:university}
\vspace{-3mm}
\end{table}

\begin{table}[!t]
\centering
\renewcommand\arraystretch{0.95}
\setlength{\tabcolsep}{15pt}
\small
\begin{tabular}{lcc}
\toprule
Models & Snips & Waihu\\\cmidrule(lr){1-3}
% \hline  %添加表格中横线
CASLU& 92.57 &69.22\\
CASLU + VAT& 92.67 &\textbf{71.75}\\
CASLU + N-Best& \textbf{93.48} & 69.90\\
\bottomrule
\end{tabular}
\label{table:ensemble}
\caption{Test accuracy (\%) of the combined methods. (Sign Test, with p-value\textless0.05)}
\vspace{-3mm}
\end{table}

\noindent \textbf{Combining with other techniques.} 
Recently, many attractive techniques of robust SLU have emerged. These techniques focus on utilizing more information, such as the classification probability distribution~\cite{ruan2020towards} or ASR N-best hypotheses \cite{li2020improving}. In \cite{ruan2020towards},  
%the authors applied virtual adversarial training (VAT) by taking the ASR hypotheses as adversarial samples.
virtual adversarial training (VAT) is applied with a Kullback–Leibler (KL) divergence term added to minimize the distance between predicted label distributions of transcriptions and ASR hypotheses to train a robust SLU model. 
%The loss function of VAT is: 
% \begin{equation}
% \begin{aligned}
% \label{eq:kl_loss}
% \mathcal{L}_{\textrm{VAT}} &= \epsilon_1 \textbf{CE}\big(\mathbf{y}, \mathbf{\hat{y}}\big)\\
% &+\epsilon_2\textbf{KL}\big(p(\mathbf{y}|X_{\textrm{trs}}^w,X_{\textrm{trs}}^p),p(\mathbf{y}|X_{\textrm{asr}}^w,X_{\textrm{asr}}^p)\big)
% \end{aligned}
% \end{equation}
In \cite{li2020improving}, 
%considering that the first best interpretation could be erroneous and noisy, 
the authors uses N-best ASR hypotheses by concatenating their texts or embeddings to improve the SLU system robustness. It is very meaningful and interesting to study whether our method can combine with them and get further gain. Therefore, we implement two combined models: CASLU+VAT and CASLU+N-best. CASLU+VAT is the integration of CASLU and VAT with 1-best hypothesis as adversarial example. CASLU+N-best exploits N-best ASR hypothesis texts and phonemes as input, contrary to just using top ASR result. 
In both settings the numbers of parameters are the same as CASLU, and the difference lies in training objectives and input data. 
%VAT adds a KL loss for prediction probabilities between transcript and ASR 1-best. N-BEST takes all ASR N-best as inputs.
Table 4 shows that both models have achieved further improvements on the two datasets, which corroborates the necessity of phoneme information and the complementarity of our method.
\begin{figure}[!t]
\begin{center}
\includegraphics[width=0.49\columnwidth, trim={10.1cm 5cm 10.9cm 4.5cm}, clip]{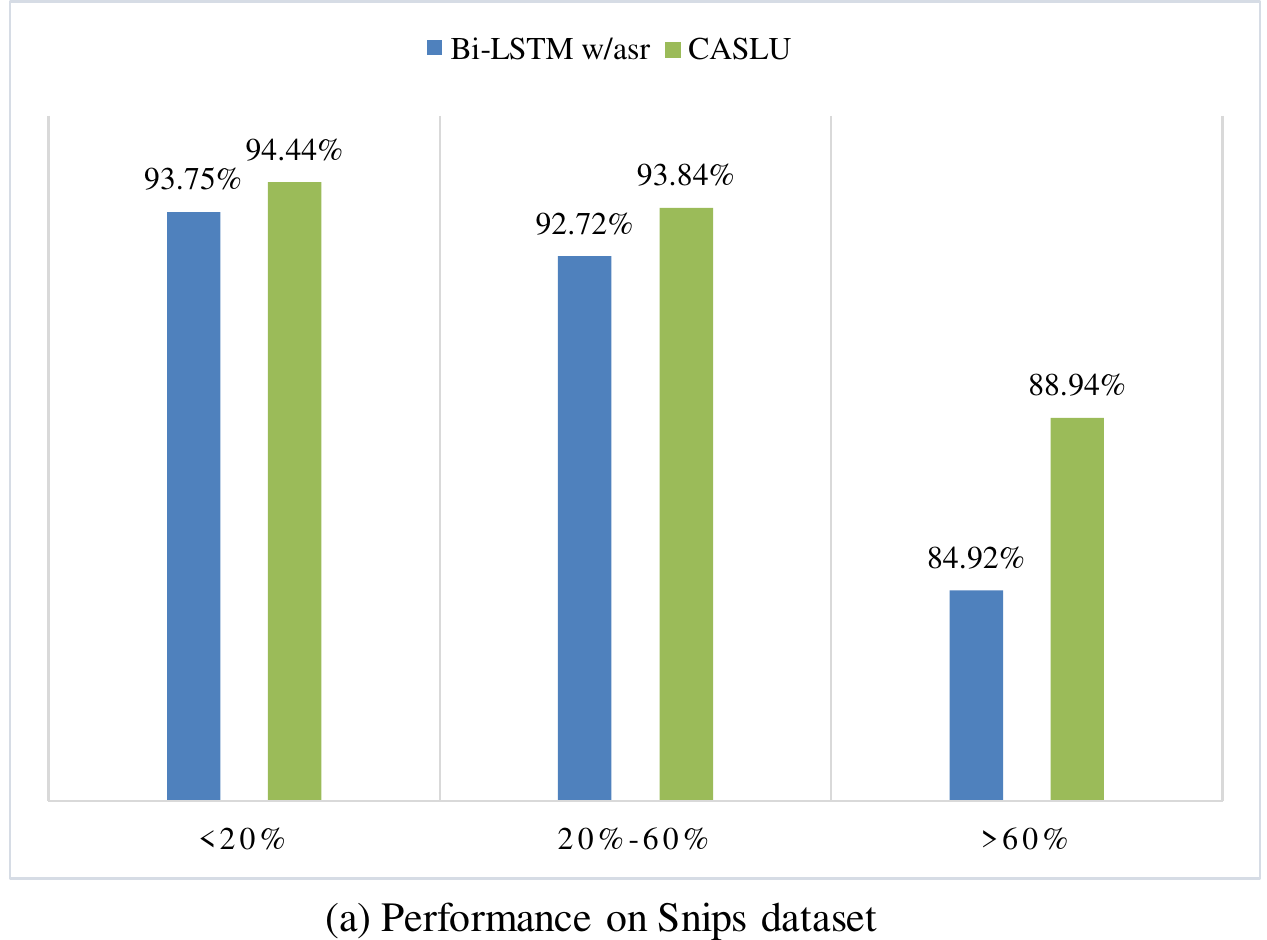}
\includegraphics[width=0.49\columnwidth, trim={8.6cm 5cm 12.4cm 4.3cm}, clip]{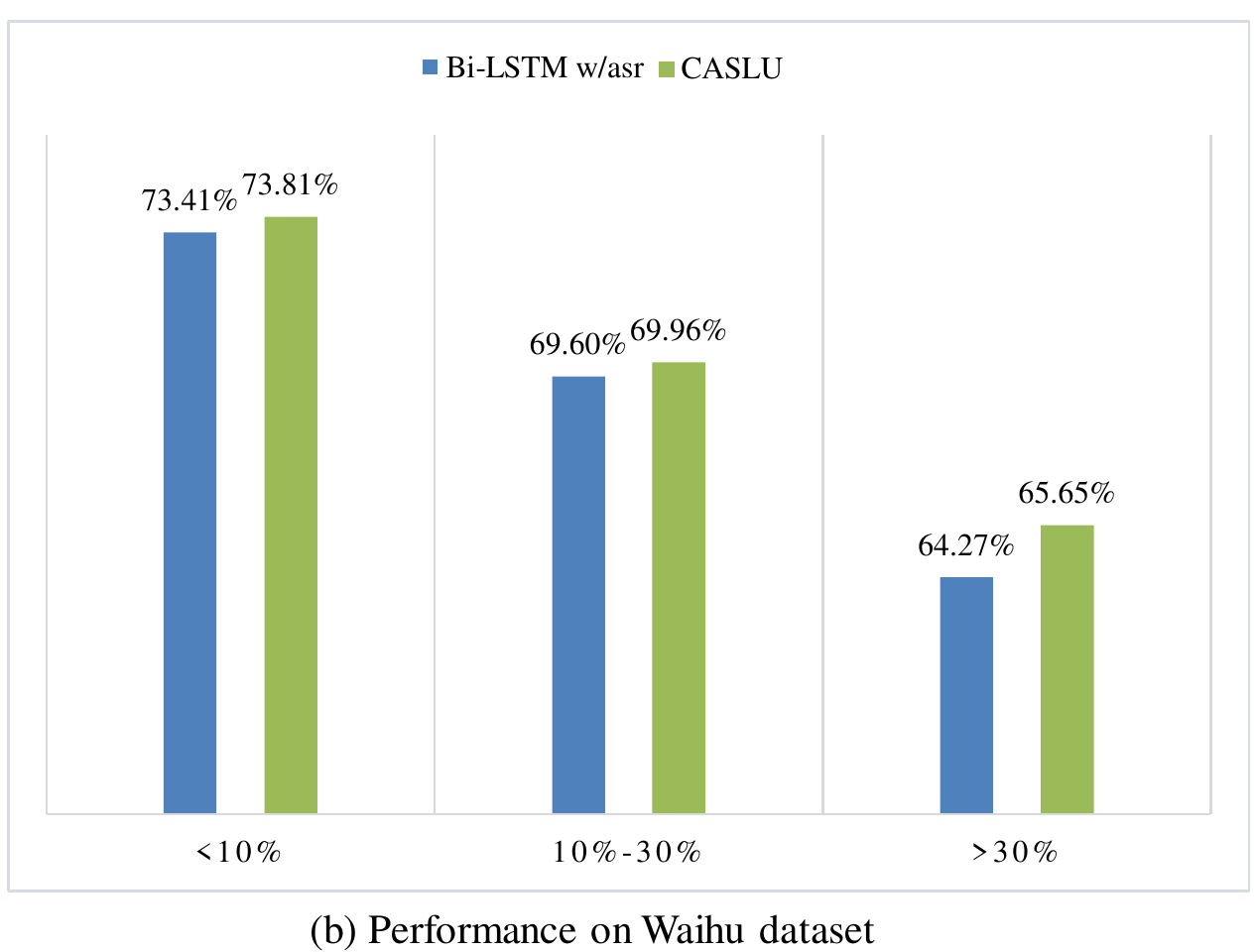}
\vspace{-2mm}
\caption{Performance comparison in accuracy of Bi-LSTM w/asr (B2) and CASLU at different WER ranges.}
\label{fig:wer_range}
\end{center}
\vspace{-3mm}
\end{figure}

\noindent \textbf{Performance at different WER ranges.}
With the fusion of phoneme and text information, it is conceivable that our model should be more robust when there are more errors in ASR hypothesis.
We verify this assumption by stratifying the Snips and Waihu test sets into three buckets based on the WER score of each instance. 
Figure~\ref{fig:wer_range} shows the results of CASLU and its counterpart without phoneme input (B2: Bi-LSTM w/ asr) on the three buckets.
We can see that not only does CASLU outperform B2 at all ranges, but the margin becomes also larger when WER increases.
For instance, CASLU is better than B2 by over four absolute points on Snips when WER is over $60\%$, which demonstrates the strong robustness of our model to ASR errors.

% To know the performance of our model under different WERs/CERs, we categorize the Snips and Waihu test sets into buckets based on the WER/CER score of each instance. From Figure \ref{fig:wer_range}, CASLU obtains more gains when WER/CER increases, which demonstrates the strong robustness of CASLU to ASR errors.

\renewcommand\arraystretch{1.5}
\begin{table}[!t]
\centering
\footnotesize
\begin{tabular}{l}
\textsl{\textbf{Transcription text:}} i want to add a song by jazz brasileiro\\ 
\textsl{\textbf{Transcription phoneme:}} ay w-aa-n-t t-uw \textit{ae-d} ah s-ao-ng\\ 
b-ay jh-ae-z b-r-ae-s-ah-l-iy-r-ow \\
\textsl{\textbf{ASR text:}} i want to i had a song by just presented right\\
\textsl{\textbf{ASR phoneme:}} ay w-aa-n-t t-uw ay \textit{hh-ae-d} ah s-ao-ng\\
b-ay jh-ah-s-t p-r-iy-z-eh-n-t-ah-d r-ay-t\\
\hline
\textsl{\textbf{Ground-Truth:}} AddToPlaylist \quad  \textsl{\textbf{Bi-LSTM w/ asr:}} PlayMusic\\
\textsl{\textbf{Multi-input:}} PlayMusic \quad \quad \quad \ \ \  \textsl{\textbf{CASLU:}} AddToPlaylist\\
\end{tabular}
\label{table:case_study}
\caption{An Example with incorrect ASR hypothesis, correctly classified using CASLU model.}
\vspace{-4mm}
\end{table}

\iffalse
\noindent \textbf{Case Study.} Table \ref{table:case_study} presents two examples in which both ASR hypotheses contain errors. 
In the left example, despite the wrong ASR text, CASLU has learnt the similarity between \textbf{ae-d} and \textbf{hh-ae-d} in the phoneme sequence, which helps our model predict the correct label.
This clearly manifests the benefit of introducing and fusing phoneme information in our model.
Contrarily, since both text and phoneme sequences of ASR hypothesis in the right example contain \textbf{UNK}\footnote{In \S\ref{sec:dataset} we use TTS and CMU lexicon tools to obtain phoneme sequence from ASR hypothesis, so it is not possible to recover UNK for phoneme.} which masks the crucial hint on the actual intent, all of the models, including our CASLU, failed in this case. 
This suggests that CASLU is limited to phoneme sequence with relatively high quality, and in order to build the model robust to even phoneme errors, the original audio also needs to be taken in to account. We leave it as our future work.
\fi

\noindent \textbf{Case Study.} Table 5 presents 
an example where the ASR hypothesis contains errors, but the similarity between \textit{ae-d} and \textit{hh-ae-d} in the phoneme sequence helps model for the correct classification. This clearly manifests the benefit of introducing and fusing phoneme information in our model.

% \begin{figure*}[!htb]
% \begin{center}
% \includegraphics[width=0.9\columnwidth, trim={10.1cm 5cm 10.9cm 4.5cm}, clip]{wer_snips.pdf}
% \includegraphics[width=0.9\columnwidth, trim={8.6cm 5cm 12.4cm 4.3cm}, clip]{wer_waihu.pdf} % \caption{\rm \centering: Accuracy comparison of the Bi-LSTM and CASLU at different WER/CER range.}
% \label{fig:wer_range}
% \end{center}
% \end{figure*}

%the following conclusions can be drawn: (1) As the WER error rate increases, the accuracy of the two methods decreases. (2) The performance degradation of our method is much lower than the baseline. These evidences show that our method has better noise immunity and, thus can address the robustness to ASR errors.

\section{Conclusions}
In this paper, we propose a novel method to enhance the robustness of SLU by fusing the information from phoneme sequence and ASR hypothesis. A fine-grained interaction module based on cross attention is devised to obtain phonetic-aware text representations and lexical-aware phoneme representations. Then two complementary representations are combined seamlessly to pursue better NLU performance. Extensive experiments were conducted to prove the effectiveness and versatility of our method. In the future, we will explore more information fusion approaches to facilitate this task.

\section{Acknowledgement}
\label{sec:acknowledgement}
This work is supported by the National Key R\&D Program of China under Grant No. 2020AAA0108600.

\bibliographystyle{IEEEbib}
\bibliography{strings,refs}

\end{document}